# Regulating Reward Training by Means of Certainty Prediction in a Neural Network-Implemented Pong Game

Matt Oberdorfer     Matt Abuzalaf

Frost Data Labs, 31910 Del Obispo, 92675, California, U.S.A.

`matt@frostdatacapital.com`     `matthew.abuzalaf@yale.edu`

## Abstract

We present the first reinforcement-learning model to self-improve its reward-modulated training implemented through a continuously improving "intuition" neural network. An agent was trained how to play the arcade video game Pong with two reward-based alternatives, one where the paddle was placed randomly during training, and a second where the paddle was simultaneously trained on three additional neural networks such that it could develop a sense of "certainty" as to how probable its own predicted paddle position will be to return the ball. If the agent was less than 95% certain to return the ball, the policy used an intuition neural network to place the paddle. We trained both architectures for an equivalent number of epochs and tested learning performance by letting the trained programs play against a near-perfect opponent. Through this, we found that the reinforcement learning model that uses an intuition neural network for placing the paddle during reward training quickly overtakes the simple architecture in its ability to outplay the near-perfect opponent, additionally outscoring that opponent by an increasingly wide margin after additional epochs of training.

## 1. Introduction

In artificial intelligence, reinforcement learning is an important field of study as it pertains to the ongoing development of "smart technology." One method by which an agent can regulate its learning is through reward modulation, wherein the agent is trained as it receives rewards for desirable actions in order to form an optimal policy [1]. An extension to this approach is that the agent can also receive a "punishment" of sorts for incorrect actions, so as to train the agent to avoid repeating the same response in future instances. Over successive iterations, the network develops a sort of predictive insight that should ultimately allow it to make a quick determination of the actions it needs to take in order to earn a reward.

We are introducing reward-modulated learning model that uses inputs and outputs of a prediction neural network to develop an "intuition" policy that is based on the results of the network's actions, and ultimately, a well-trained agent can use this intuition to make predictions even in a complex environment, where simple pattern recognition may not capture all possible intricacies.

One key to the successful training of any reinforcement-learning model is an ability to make quick and accurate predictions about the actions it needs to take to be rewarded. If the agent can be trained to make a good prediction in the context of its environment, then it can be further developed to accomplish particular goals in that setting. The obstacle to overcome in the implementation of such an approach, however, tends to be the difficulty in setting up a neural network that can learn efficiently and accurately to predict successful behavior in a previously unknown environment.

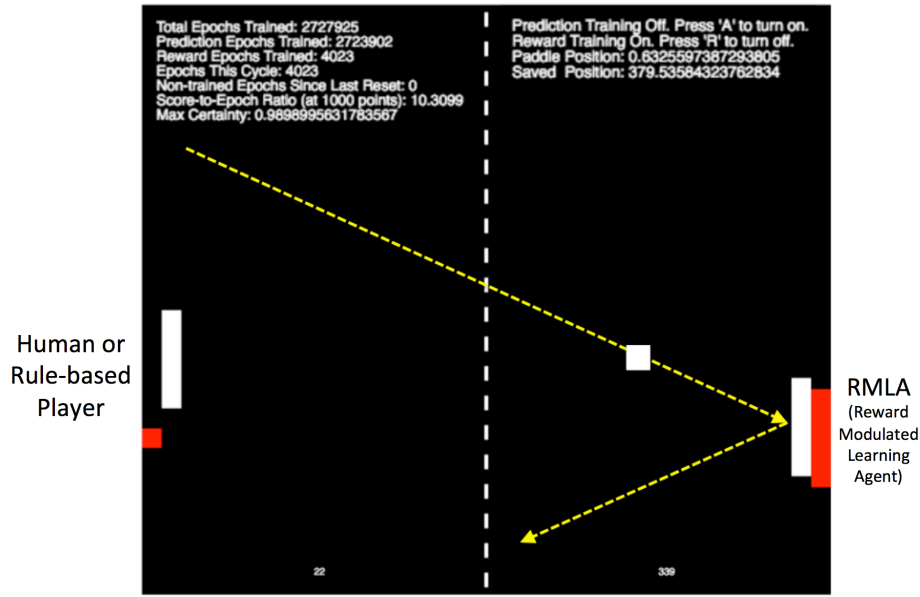

*Figure 1*

At the most basic level, success is constituted here by using such inputs for the network that will capture the fullness of the problem and make a prediction that reflects this complexity. Once this is accounted for, more features can be added, such as additional rewards that are weighted based on their relative importance to the problem, switches that allow for alternation between training styles, and even the implementation of additional learning networks that run in parallel with the original.

To test how our learning model may capture and learn a complex problem, the arcade game Pong was selected as a good candidate. It has been shown [2] that arcade games can represent both a challenge problem and provide a platform and methodology for evaluating the development of general, domain-independent AI technology.

As one of the earliest arcade games, the program's environment is not particularly complicated: at its basis, the environment involves passing a ball back and forth between paddles as one would do in a game of ping-pong (Figure 1). At the same time, however, the game reflects a degree of complexity in terms of the number of ways any particular return can end up going. The ball can bounce off the top and bottom walls. It can reflect off the paddle at a variety of different angles. And if the ball hits the corner of the paddle, then the return will be made at a greater speed. These factors make for a game where it may not immediately be so simple to predict where to place the paddle for a successful return, and as such, this complex environment is an excellent candidate for an agent to learn by reward-modulated training.

We made use of two reward-modulated neural network architectures of different complexities, differing primarily in their methods of placing the agent paddle while the program was training. The first architecture, referred to as the "simple reward" architecture, simply placed the Reward-Modulated Learning Agent (RMLA) paddle in a random position every time the ball was returned in its direction, only then training when a successful hit was made.

The second architecture, referred to as the "four-network" architecture, was more complex, using the same-reward modulation as before while at the same time implementing three additional neural networks in order to allow the RMLA to learn a "certainty" about whether or not its paddle placement was correct. This allowed for the use of the fourth "intuition" network as a policy to expedite the training process by minimizing the need for random paddle placement during reward training. The certainty and intuition networks enable the learning model to assess the state and derive actions similar results as Q-learning [3] while entirely based on back propagation networks.

## 2. Related Work

Reinforcement learning is an area of machine learning wherein a machine learns how to perform optimally by its own trial-and-error processes, rather than by external human correction. In reinforcement learning, the selection of correct decisions is made so as to maximize preordained "rewards" in the long term, and the network is strengthened upon reception of these rewards so that it performs similarly in future instances. [1] In general, this process works as follows: over a series of stages or events, the learning agent will make some arbitrary number of decisions or interactions. The policy of the RMLA guides its behavior at any given time in the process (which can be random or deterministic). Positive actions or stages in the process are assigned some numeric representation of desirability by means of a reward function, and it is the RMLA's ultimate goal to maximize its reward in the long run, thus essentially "learning" positive behaviors within the context of its environment along the way.

There are several methods by which reinforcement can occur. One traditional method is learning from delayed rewards [4], a process by which the program uses temporal differences over successive estimates to try to push closer to an optimal policy [5].

An alternative approach is using reward-modulation to train its neural network. In this method, the network is engaged and trained when the agent carries out a desirable action, which, is pre-designated by the reward function [6]. As the network is trained, it learns to contextualize the actions that earned it the reward as a functional output of its variable inputs, therefore being able to replicate the action that led to a reward when a similar situation presents itself in future instances. A practical comparison of confidence estimation methods for neural network architectures examined their effectiveness [7]. In 2013, a deep Q-learning variation was used to create an agent that learns how to play a variety of Atari games [8] and outperform human players.

## 3. Regulating Reward Training by Means of Certainty Prediction

In our approach, we used multiple layers of representation [9] combined with sequenced neural networks [10] to develop a policy that consists of backpropagation networks. We used extensive reward modulation in training, and the key question asked was whether additional functionality could be implemented in order to help the reward network learn how to play faster than it would simply through the brute-force method of placing the paddle randomly.

The result of this inquisition was the creation of the "four-network" architecture, wherein the placement of the paddle during reward training was regulated by three additional networks, which would learn to reliably estimate a confidence for the likelihood that the agent paddle would successfully return the ball and subsequently use this estimation to conclude where to place the paddle during training. The Hypothetically, this would result in a more efficient training session and a faster convergence of the reward network's paddle position prediction to the correct value in every epoch.

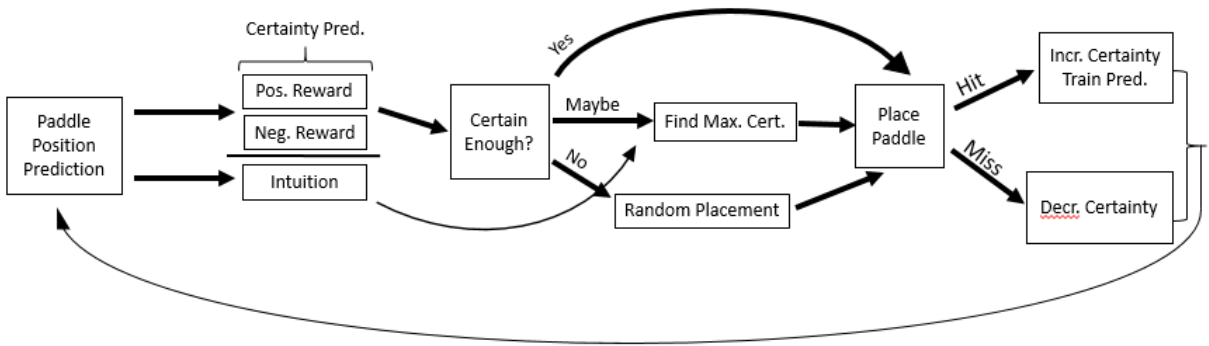

*Figure 2*

The arcade game used in this experiment was a Pong implementation featuring the basic functionality of the original Atari Pong game, including two moving paddles, a ball that could bounce off of the paddles, as well as the top and bottom of the screen, a "serve" function that would serve the ball in a random direction after a score, and a "corner hit" feature that would bounce the ball at a faster speed off the corner of the paddle. In this version, the agent paddle would track the movement of the ball with a small amount of lag such that it was good at defense but also imperfect at the same time. Basic additions to the program included a score-tracking function, a switch that would place the player paddle under near-perfect computer control (the paddle was made to "wobble" in place so that it would miss occasionally), and a "speed-up" button that increased the speed of gameplay so that thousands of epochs of training could occur quickly.

The simple reward network uses a 6:10:1 architecture to learn how to predict paddle position based on six inputs, the y-position of the ball relative to the top of the screen, the y-position of the ball relative to the bottom of the screen, the y-velocity of the ball if positive, the y-velocity of the ball if negative, the x-velocity of the ball, and the angle phi the ball makes with respect to the horizontal, all normalized so that they take on values between 0 and 1. The training is reward-modulated, and a reward function is set such that back-propagation occurs whenever the agent paddle successfully makes contact with and returns the ball. A switch is programmed to turn "reward training" on and off, and when training is active, the agent paddle is placed randomly. The program is only rewarded when the random placement yields a successful return.

For the four-network reward architecture, three additional networks are added to supplement the previous prediction network, which is now constructed with a 6:12:1 architecture. Two of the three new networks are the "Positive Reward" and "Negative Reward" networks with 8:16:2 architectures. The inputs are the same as before but are supplemented with the normalized paddle position prediction defined as the output of the original network, along with that prediction value subtracted from 1. The outputs of these networks involve the position prediction as before in addition to a binary variable. For the Positive Reward network, it takes on a value of 1 for a hit and 0 for a miss. For the Negative Reward network, the opposite is true- the binary variable registers a 0 for a hit and a 1 for a miss.

The final network in the four-network program is the "Intuition Network," which uses an 8:24:3 architecture (the outputs are the position prediction and two binary variables responding to a hit and a miss).

The four-network architecture's policy and structure follows the flow displayed in Figure 2 at the bottom of this document. To summarize: the Prediction network feeds into Positive Reward, Negative Reward, and Intuition networks, which run in parallel and whose outputs determine the placement of the paddle when reward training is switched on.

The binary outputs of the Positive Reward and Negative Reward networks are used in the construction of a "certainty" prediction, which involves the product of the activated positive reward output with the difference of the activated negative reward output from 1. If this certainty value is above a certain threshold (95% in this case), and if the mean of the last 10 certainties is also above a certain threshold (80% in this case), then the paddle will be placed on the basis of the trained prediction, rather than randomly. If the value is below this threshold but above a certain lower threshold, the Intuition Network is invoked in order to place the paddle based on the position where certainty is estimated to be maximized. And if the certainty is below even the low threshold, the paddle will be placed randomly, as it is then theoretically more likely to yield a successful return than would a confidently wrong placement.

We create a series of experiments to validate the hypothesis that this adaptive reward-modulated training is more efficient at correct placement than is placing the paddle in a random position for every training epoch.

## 4. Results of Experiments

One of the two goals was simply to check whether or not the reward-modulated learning was successful at training the agent to play Pong. The second goal was to compare the two programs, the "Simple Reward" and "Four-Network" architectures, in order to see whether the introduction of confidence-regulated reward training could result in a significantly more efficient training process.

To test, the two games were run with reward training switched on, and agent performance was checked at certain intervals of training epochs: 500K, 1M, 2M, and 5M. At these many epochs, reward training was switched off and the agent was allowed to perform on its own. Once the trained agent had scored 10,000 points with training off, the "human" score was recorded, and the reward training continued. The relative success and performance of the two systems could then be recorded and compared.

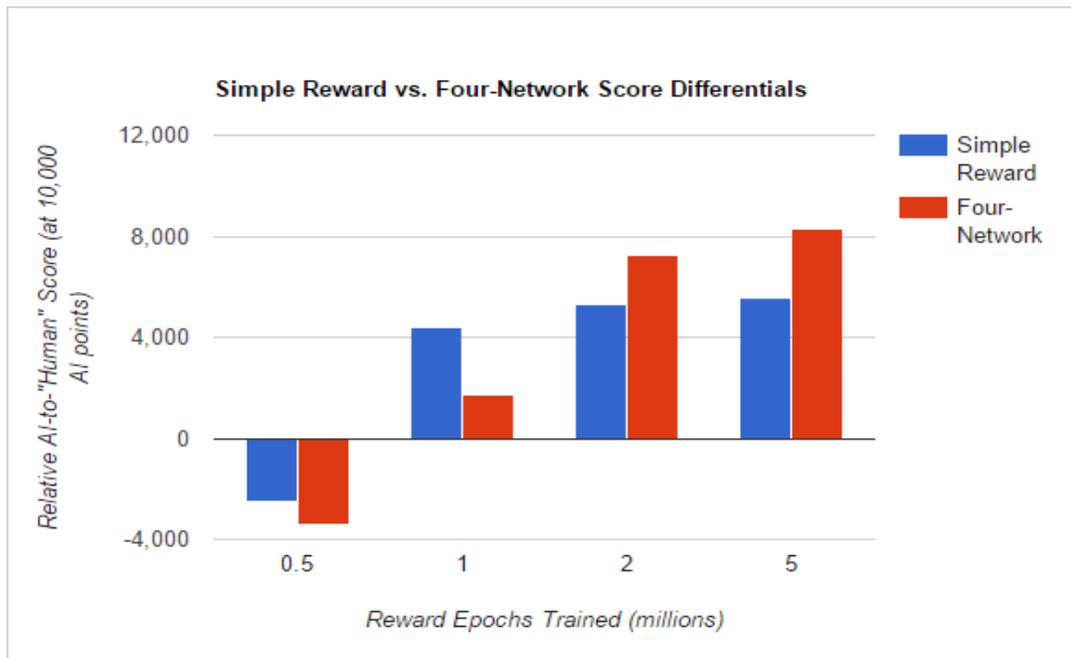

*Figure 3*

*Table 1. "Perfect Human Player" Scores at agent Score of 10,000 After Specified Number of Reward Training Epochs*

|  | Simple Reward | Four-Network |
|---|---|---|
| 500K epochs | 12479 | 13446 |
| 1M epochs | 5616 | 8289 |
| 2M epochs | 4696 | 2752 |
| 5M epochs | 4454 | 1708 |

As represented by Figure 3 and Table 1 above, both reward networks were able to outscore the near-perfect "human" player (the computerized perfect ball tracker with a slight wobble) at some point between 500,000 and 1,000,000 epochs of training, and as training continued, both trained agent programs continued to improve their margins of victory over the near-perfect player. Theoretically, continued training for millions of additional epochs would allow for the continued betterment of the agent programs until they approach "perfect" predictive performance.

In comparing the two programs, we validated that the more developed four-network architecture yields better results. Namely, at the same number of reward training intervals, the four-network program begins to outperform the simple reward program by an increasingly large margin, though both programs manage to learn how to outscore the human player. In other words, the four-network program should of the two choices theoretically develop perfect predictive capability more quickly.

## 5. Analysis and Evaluation

To weigh in on the first question specified in the results section, it became evident that the reward-modulated training methods applied here were successful. If success is qualified as the reliability of the RMLA in it being able to outscore a particularly good "human player," then the result is an astounding success for both the simple reward and four-network programs.

Both begin to outscore the "human" at some point between 500K and 1M epochs of training, which, with the speed-up function programmed into the script, can be accomplished in under five minutes. At this point, an actual imperfect human playing against the agent would be soundly defeated time and time again. With additional training, both programs begin to win by greater margins and become essentially perfect within a relatively short amount of time. This reflects well on the choice of variable inputs and architectures for the two networks, as these are the key elements in the agent's success and efficiency.

As for the comparison of the two approaches' relative performances, it also became evident that the four-network architecture is by far the more efficient of the two. Looking at Figure 2 and Table 1, it is evident that while the simple reward network may appear to perform better initially (this may simply be the result of variability from the small sample size of epochs trained), the four-network program demonstrably has the better performance as the training goes on.

More importantly, its margin of improvement between intervals is larger than that for the simple reward program. In other words, it gets better faster. At its relative rate of improvement, it can be projected that the four-network program will converge to perfect performance much more quickly than would the simple reward program, and since efficiency is key in the operation of neural networks, the fact that this method can adaptively regulate and expedite the reward training process is particularly noteworthy.

## 6. Conclusion

We successfully validated the hypothesis that our reward-modulated learning model outperforms near-perfect machine and human opponents playing the arcade game pong. The reward modulation model and policy consisted of four sequenced neural networks whereby the second and third network predicted the certainty of receiving a reward that was predicted by the first network. A fourth neural network was trained to generate an action in case the certainty was low. Even with a simple reward network that involved the use of nothing more than random paddle placement to train the agent on successful returns, the program was able to learn how to outscore its opponent in less than a million training epochs, which translates to less than five minutes of training in real time. Additionally, performance only grew better with additional training, and it seems that in a reasonably short amount of time, the simple reward network would have trained a near-perfect agent. The particularly noteworthy conclusion here, pertains to the further success of the reward modulated machine learning model in adapting and optimizing the reward training process such that it allowed the agent to increase performance at a faster rate, converging towards perfection much more efficiently than other reward architectures, which appears to have much more pronounced diminishing returns on improvement.

## 7. Suggestions for Further Research

A comparison between this and other methods of training an agent to play arcade games, such as deep Q-learning [8], will help to validate which model shows better training convergence and task accomplishments. In future experiments, we will examine whether the four-network method developed here could be generalized and implemented for other challenges including other arcade games. If in an equivalent number of training epochs our approach of reward-modulated program could outperform Q-learning based algorithms, then the four-network method developed here would be an alternative machine learning method for use in more general applications than the one featured in this experiment.